\documentclass[conference]{IEEEtran}
\IEEEoverridecommandlockouts
\usepackage{amsmath,amssymb,amsfonts}
\usepackage{algorithmic}
\usepackage{graphicx}
\usepackage{textcomp}
\usepackage{xcolor}
\usepackage{textgreek}
\usepackage{paralist}
\usepackage[style=ieee, backend=biber]{biblatex}
\def\BibTeX{{\rm B\kern-.05em{\sc i\kern-.025em b}\kern-.08em
    T\kern-.1667em\lower.7ex\hbox{E}\kern-.125emX}}
\begin{document}

\title{Bounding the Black Box: A Statistical Certification Framework for AI Risk Regulation\\
}

 \author{\IEEEauthorblockN{Natan Levy}
 \IEEEauthorblockA{\textit{School of Computer Science and Engineering} \\
 \textit{The Hebrew University of Jerusalem(HUJI)}\\
 Jerusalem, Israel \\
 Natan.Levy1@mail.huji.ac.il}
 \and
 \IEEEauthorblockN{Gadi Perl}
 \IEEEauthorblockA{\textit{dept. name of organization (of Aff.)} \\
 \textit{The Hebrew University of Jerusalem(HUJI)}\\
 Jerusalem, Israel \\
 Gadi.perl@mail.huji.ac.il}
 }

\maketitle

\begin{abstract}
Artificial intelligence now decides who receives a loan, who is 
flagged for criminal investigation, and whether an autonomous 
vehicle brakes in time. Governments have responded: the EU AI 
Act, the NIST Risk Management Framework, and the Council of 
Europe Convention all demand that high-risk systems demonstrate 
safety before deployment. Yet beneath this regulatory consensus 
lies a critical vacuum: none specifies what "acceptable risk" 
means in quantitative terms, and none provides a technical method 
for verifying that a deployed system actually meets such a 
threshold. The regulatory architecture is in place; the 
verification instrument is not.

This gap is not theoretical. As the EU AI Act moves into full 
enforcement, developers face mandatory conformity assessments 
without established methodologies for producing quantitative 
safety evidence - and the systems most in need of oversight are 
opaque statistical inference engines that resist white-box 
scrutiny.

This paper provides the missing instrument. Drawing on the 
aviation certification paradigm, we propose a two-stage framework 
that transforms AI risk regulation into engineering practice. In 
Stage One, a competent authority formally fixes an acceptable 
failure probability $\delta$ and an operational input domain 
$\varepsilon$ - a normative act with direct civil liability 
implications. In Stage Two, the RoMA and gRoMA statistical 
verification tools compute a definitive, auditable upper bound on 
the system's true failure rate, requiring no access to model 
internals and scaling to arbitrary architectures. We demonstrate 
how this certificate satisfies existing regulatory obligations, 
shifts accountability upstream to developers, and integrates with 
the legal frameworks that exist today.
\end{abstract}



\section{Introduction}
\label{sec1:Inrtoduction}

Artificial intelligence systems now operate across domains that were once the exclusive province of human judgment, in areas from medical diagnostics and judicial decision-making to employment screening and critical infrastructure. As deployment has expanded, so has regulatory ambition. Jurisdictions worldwide have converged on a common model: risk-based regulation that classifies AI systems by their potential for harm and imposes proportionate obligations on their developers and deployers.

This convergence is significant. The EU AI Act \cite{euaiact2024}, the NIST AI Risk Management Framework \cite{nist2023rmf}, and China's suite of sectoral regulations \cite{li2024china} all rest on the same foundational premise: that a system's risk can be assessed before deployment, and that compliance with an acceptable risk standard can be verified. What these frameworks have not done is specify what "acceptable risk" means in quantitative terms, or provide technical methods for establishing whether a given system actually meets such a threshold. The regulatory architecture is in place; the measuring instrument is missing.

This gap is not theoretical. Recent empirical work shows that leading large language models shift between refusing and complying with harmful requests at rates of 5–20 percent in response to minor syntactic perturbations \cite{stanovsky2025}. If benchmark performance is this sensitive to surface-level variation, regulatory approval based on benchmarks will, as Stanovsky et al. conclude, "falsely certify unsafe systems."~\cite{stanovsky2025}. This problem is structural: neural networks are statistical inference systems without stable world models, and their behavior across the unbounded space of real deployment conditions cannot be guaranteed by performance on any finite test set.

Legal scholars have reached the same conclusion from a policy perspective. Ebers \cite{ebers2024} shows that the EU AI Act's risk tiers are predefined and rigid, with providers largely self-certifying compliance. Kaminski \cite{kaminski2023} warns that governance frameworks built around categorical risk labels tend to entrench techno-correctionism rather than establish structural accountability. Risk-based regulation without a measurable definition of risk is incomplete.

To bridge this regulatory gap, this article proposes a comprehensive framework designed to complement existing legislation with mathematical tools capable of statistically verifying compliance against predefined risk thresholds. The underlying verification instruments are RoMA (\textit{Robustness Measurement and Assessment}) and its group-aware extension gRoMA~\cite{LeKa21,LeYeKa23,LeAsKa25} --- established statistical methods for certifying the robustness of neural networks against bounded perturbations; the contribution of this paper is not a new algorithm but the regulatory application framework that takes these existing theoretical instruments and translates their outputs into a pre-deployment compliance certificate with determinate legal and normative meaning within existing AI governance structures. The specific contributions of this paper are: 
\begin{inparaenum}[(i)] 
\item a formal two-stage certification architecture that decouples the normative determination of acceptable failure probability~$\delta$ from its technical verification, eliminating the conflation of value judgment and engineering assessment that characterises current conformity assessment practice; 
\item an interface specification between the two stages that gives the certificate determinate pass/fail semantics --- the RoMA test is meaningful only relative to a publicly stated~$\delta$, and a re-test against the same~$\delta$ and~$\varepsilon$ either confirms or defeats the original certificate; 
\item a demonstration that the procedure is deployable on an industrially relevant safety-critical architecture without access to model internals, presented as a structured proof-of-concept on a high-resolution autonomous braking system; and \item a mapping of the two-stage structure onto existing compliance obligations under the EU AI Act and NIST AI RMF~\cite{euaiact2024,nist2023rmf}, grounding the engineering design in the regulatory requirements it is intended to satisfy. 
\end{inparaenum}

The remainder of this paper is organized as follows: Section \ref{sec2:RegulatoryLandscape} examines the comparative regulatory landscape across the European Union, China, and the United States, delineating the shared structural assumptions underlying these governance models. Section \ref{sec3:EvaluatingSystemSafety} maps the current state of system validation methodologies and their technical boundaries. Section \ref{sec4:Framework} introduces the proposed two-stage framework, detailing both the quantitative definition of acceptable risk and the exact mechanisms for its verification. Section \ref{sec5:caseStudy} operationalizes this methodology through a targeted case study, demonstrating the framework's deployment in a safety-critical context. Section \ref{sec6:ThreatsToValidity} critically evaluates the methodological limitations and threats to the framework's validity. Finally, Section \ref{sec7:conclusion} concludes the study and identifies promising directions for future research.

\section{The Regulatory Landscape}
\label{sec2:RegulatoryLandscape}
Risk-based regulation is now the dominant approach to AI governance worldwide. Jurisdictions have converged on the principle that regulatory obligations should be proportionate to potential harm, but they differ substantially in their normative foundations and enforcement architectures. Understanding these differences is important for one central reason: no major framework has yet specified what ``acceptable risk'' means in measurable terms, nor how compliance with such a threshold can be technically verified.

\subsection{European Union}
The EU AI Act~\cite{euaiact2024} is the world's first comprehensive binding AI law. It classifies systems into four tiers: prohibited practices, high risk, limited/transparency risk, and minimal risk. High-risk systems, covering domains such as employment, law enforcement, and judicial administration, must pass conformity assessments, implement human oversight, and maintain extensive technical documentation before market entry. General-purpose AI models face additional obligations, with systemic-risk requirements triggered above a $10^{25}$ FLOPs compute threshold~\cite{euaiact2024, ec2025gpai}. Implementation is actively being phased through 2026.

Legal scholarship has focused directly on this operationalization gap. Ebers~\cite{ebers2024} argues that the risk categories are predefined and inflexible, and that providers largely self-certify without independent, technical scrutiny. Wachter~\cite{wachter2024} highlights how industry lobbying produced broad carve-outs under Article 6(3). Furthermore, Laux, Wachter, and Mittelstadt~\cite{laux2024} note that the Act conflates risk acceptability with trustworthiness, failing to capture the institutional legitimacy the latter requires.

\subsection{China}
In contrast, China has adopted an incremental, sector-specific approach through binding administrative regulations rather than a single unifying statute. Three measures form the core framework:
\begin{inparaenum}[(i)]
\item The Provisions on Algorithmic Recommendations (2022), requiring CAC algorithm filing and prohibiting opinion manipulation~\cite{cac2022algo}; 
\item The Provisions on Deep Synthesis (2023), mandating real-name verification and the labeling of AI-generated content~\cite{cac2023synthesis}; and 
\item The Interim Measures for Generative AI Services (2023), requiring security assessments and adherence to ``core socialist values'' for services with ``public opinion attributes''~\cite{cac2023genai}.
\end{inparaenum}
Here, risk classification hinges primarily on a system's capacity to shape political discourse rather than its potential harm to individual rights, a fundamentally different normative foundation from the European model.

\subsection{The United States}
The United States has historically relied on voluntary frameworks and executive action rather than binding federal legislation. The NIST AI RMF~\cite{nist2023rmf} organizes risk governance around four core functions, Govern, Map, Measure, and Manage, and has become a \textit{de facto} global reference standard. Its Generative AI Profile~\cite{nist2024genai} identifies thirteen GenAI-specific risks and over 400 suggested actions. Executive Order 14110 (2023), which initially imposed safety-testing requirements on prominent AI developers, was revoked by Executive Order 14179 (2025), explicitly shifting federal policy toward economic competitiveness and innovation over strict risk mitigation~\cite{trump2025eo}. As of early 2026, no binding federal AI legislation has been enacted.

\subsection {Common Assumptions and Their Limits}
\label{secIID:CommonAssumption}
Despite their divergent foundations, all three frameworks share two structural assumptions:
\begin{inparaenum}[(i)]
\item that a system's risk can be reliably assessed prior to deployment (ex-ante classification); and 
\item that standardized benchmarks can reliably verify system safety (benchmark reliance). 
\end{inparaenum}
Both assumptions remain technically ungrounded. Stanovsky et al.~\cite{stanovsky2025} demonstrate that minor syntactic perturbations, altering prompt formatting without changing semantic meaning, cause leading Large Language Models to shift between refusing and complying with harmful requests at rates of 5--20 percent. This definitively proves that benchmark-based certification can falsely certify unsafe systems. Neural networks operate as statistical inference engines without stable, underlying world models; therefore, benchmark scores cannot safely predict behavior across the infinite space of real-world deployment inputs.

\subsection{Current Regulatory Crisis}
The Council of Europe Framework Convention on AI (CETS No. 225), entering into force in November 2025, serves as the first binding international AI treaty~\cite{coe2024}. Concurrently, South Korea's AI Basic Act (effective January 2026) and the EU Digital Omnibus~\cite{eu2025omnibus} confirm the global consolidation of risk-based governance. These overlapping developments crystallize the core problem: while risk-based regulation is now an international legal norm, no major framework specifies what level of risk is acceptable or how compliance can be mathematically verified. Filling that critical gap is the objective of the framework proposed below.

\subsection{The Aviation Paradigm: Certifying Traditional Software}
The absence of verifiable risk thresholds in current AI governance stands in stark contrast to mature, safety-critical domains. In traditional software engineering, particularly within aerospace, established certification standards provide exact methodologies for ensuring system safety. Unlike the statistical inference models of modern neural networks, deterministic code can be systematically validated against strictly defined operational requirements~\cite{LaNi11}.

Today, the aviation industry relies on a universally recognized methodology that creates a level of assurance so dependable that society inherently trusts these systems to fly safely~\cite{rushby2011new}. This certification standard is the culmination of decades of expert consensus and empirical refinement. A cornerstone of this framework is DO-178~\cite{FAA93}, which mandates an uncompromising, fully traceable lifecycle for airborne software development. By requiring a measurable demonstration that a system will perform its intended function under all foreseeable conditions, DO-178 has achieved an extraordinary safety record; to date, there are no recorded commercial aircraft crashes attributed to a failure in software developed and certified under this standard~\cite{rushby2011new}.

The software-level assurances of DO-178 are directly derived from the broader system-level requirements established by ARP 4754~\cite{LaNi11}. This foundational guideline systematically analyzes every functionality within the aircraft to determine its criticality and dictates the acceptable failure probability with which the system must comply. For example, a safety-critical functionality, such as flight control, must meet a quantitative failure probability of no more than $10^{-9}$ per working hour. This methodology proves that verifiable, quantitative safety thresholds are structurally possible for highly complex systems, highlighting precisely the measuring instrument that is currently missing from AI regulation.

\subsection{Early Attempts at Neural Network Certification}
Despite the strict requirements of traditional software certification, the aerospace industry increasingly recognizes the transformative value that artificial intelligence, specifically neural networks, can bring to aviation. Consequently, the focus has shifted toward integrating these advanced capabilities without compromising established safety margins. To this end, EUROCAE, functioning as a key standard-developing organization in coordination with the European Union Aviation Safety Agency (EASA), established a dedicated working group to chart a viable pathway for the verification and certification of neural networks in safety-critical airborne systems~\cite{gabreau2022toward}.

Early efforts to bridge this certification gap have predominantly centered on formal verification techniques. Foundational research by Katz et al. \cite{KaBaDiJuKo21} demonstrated that a deep neural network, specifically an implementation for the Airborne Collision Avoidance System (ACAS Xu), could be formally verified against exact safety properties using Satisfiability Modulo Theories (SMT) solvers~\cite{KaBaDiJuKo21}. However, this mathematically precise approach faces two critical limitations. First, the underlying computational problem of formally verifying neural network properties is NP-complete~\cite{KaBaDiJuKo21}, rendering these methods computationally intractable for networks exceeding a few thousand neurons and precluding their application to modern, over-parameterized architectures. Second, SMT-based verification inherently requires complete white-box access to the model's architecture and weights~\cite{LeKa21}. This structural prerequisite presents an insurmountable barrier for the evaluation of proprietary, commercial models, which are overwhelmingly deployed as opaque, black-box systems.

\subsection{The Shift to Statistical Verification}
To overcome the computational and structural limitations of formal methods, recent research has shifted toward statistical verification~\cite{HuHuHuPe21,CoRoZi19,WeRaTeYe18,KaLeRaYe24}. Instead of seeking deterministic proofs, this approach evaluates whether a neural network complies with a predefined risk or safety threshold within a calculable confidence interval. Statistical verification offers two critical advantages:
\begin{inparaenum}[(i)]
\item It operates effectively on opaque, black-box architectures; and 
\item provided an appropriate sampling methodology is utilized, it scales to evaluate modern, highly parameterized models~\cite{LeKa21}. 
\end{inparaenum}
The primary drawback, however, is inherent to its probabilistic nature: it lacks exact deterministic precision, and its reliability is fundamentally dependent on the quality and distribution of the underlying sampling method~\cite{LeKa21}.

\section{Evaluating Systemic Safety: The Mechanics of RoMA and gRoMA}
\label{sec3:EvaluatingSystemSafety}

To operationalize statistical verification for regulatory compliance, certification frameworks require tools capable of assessing models at scale without demanding white-box access. The Robustness Measurement and Assessment (RoMA) method, alongside its global extension gRoMA, provides a concrete mechanism for calculating these precise probabilistic assurances~\cite{LeKa21}.

\subsection{Local Robustness via RoMA}

The RoMA algorithm evaluates the local robustness of a deep neural network by calculating the probability that a random input perturbation will result in a distinct misclassification. This approach aligns directly with the fault-tolerance paradigms used in aviation, which model failures as natural, random occurrences rather than the product of adversarial malice~\cite{LaNi11}.

RoMA operates entirely on black-box architectures, requiring no prior assumptions regarding the network's topology or internal activation functions. The methodology proceeds through the following structured steps:

\begin{itemize}
    \item Sampling: The algorithm generates a specified number of randomly perturbed samples around a fixed input point, ensuring the perturbations do not exceed a maximum defined size, denoted as $\epsilon$.
    
    \item Data Extraction: Each sample is passed through the neural network to retrieve its output confidence vector, from which RoMA extracts the ``highest incorrect confidence'' score.
    
    \item Distribution Normalization: RoMA tests the extracted scores using a goodness-of-fit procedure, such as the Anderson-Darling test~\cite{An11}, to evaluate if they follow a normal distribution. If the data is abnormally distributed, the algorithm applies a Box-Cox power transformation~\cite{BoCo82} to normalize the measurement scale.
    
    \item \textbf{Probabilistic Assessment:} Once normalized, the algorithm uses standard Z-score calculations and the Gaussian cumulative distribution function to evaluate the exact probability that a random perturbation will cross a specified confidence threshold, $\delta$, resulting in an adversarial failure~\cite{LeKa21}.
\end{itemize}

It is important to note that the original authors of RoMA conceptualize the algorithm strictly as a safety certification tool, drawing direct parallels to aviation standards like DO-178. Consequently, the methodology is exclusively calibrated to evaluate internal failure modes-errors originating from the model's own architecture and statistical representations. The framework makes no claim to quantify vulnerabilities caused by external harm, such as malicious cybersecurity exploits or coordinated adversarial attacks. This deliberate boundary ensures that the resulting certificate measures the system's baseline operational reliability, rather than its resilience against hostile external actors.

\subsection{Scaling to Global Categorial Robustness via gRoMA}
\label{seIIB:Scaling}

While RoMA provides exact statistical guarantees for individual data points, system-level certification requires an understanding of how a model behaves across its entire operational domain. The gRoMA (global RoMA) tool scales the RoMA methodology to measure the probabilistic global categorial robustness of a neural network~\cite{LeYeKa23}.

Rather than assuming uniform robustness across a model, gRoMA isolates the analysis to specific output categories, acknowledging that a network may exhibit varying levels of reliability depending on the classification task. The tool executes this global assessment as follows:

\begin{itemize}
    \item Representative Selection: gRoMA randomly draws a random finite set of inputs that represent a specific output category of interest from the input space.
    
    \item Iterative Local Assessment: For every correctly classified sample within this drawn set, the tool invokes the underlying RoMA algorithm to calculate an individual probabilistic local robustness score.
    
    \item Statistical Aggregation: The individual scores are mathematically aggregated, commonly through a numerical average, to establish a comprehensive global robustness score for that specific category.
    
    \item Error Bounding: To meet the strict assurance requirements of safety-critical deployments, gRoMA utilizes Hoeffding's inequality to compute a formal error bound~\cite{ho63}. This guarantees that the estimated global robustness score deviates from the true population value by no more than a tightly controlled, predefined margin.
\end{itemize}

By quantifying both local perturbation risks and global category-specific vulnerabilities, these tools provide the measurable thresholds currently missing from top-down AI governance frameworks.

\subsection{Empirical Validation of RoMA}
Formal verification methods offer the ultimate mathematical guarantee by determining the precise frequency of adversarial perturbations within a specified domain~\cite{KaBaDiJuKo21}. To establish the reliability of statistical approximation for critical system integration, recent empirical work validated the RoMA framework directly against these mathematically exact baselines~\cite{LeAsKa25}. In this comparative study, researchers established ground-truth robustness probabilities using the Exact Count algorithm. This formal methodology computes the true violation rate by systematically and recursively partitioning the input space into definitive sub-regions, allowing for an exact calculation of the probability of encountering an adversarial perturbation. 

Because this exhaustive space-partitioning suffers from exponential computational complexity, calculating the definitive ground truth is strictly constrained to small-scale neural networks, such as the 300-neuron ACAS Xu models utilized in aviation collision avoidance. Statistical methodologies, however, are not constrained by this architectural bottleneck and can scale effectively~\cite{KaBaDiJuKo21}. When evaluating these networks where exact formal computation remains tractable, the results demonstrated exceptional alignment: the statistical probabilities generated by RoMA deviated from the true formal probabilities by less than 1\%. Furthermore, while the Exact Count algorithm required hours of computation-and consistently timed out after 24 hours on slightly larger benchmarks --- the statistical framework delivered these highly precise estimates in under 16 minutes. This sub-1\% error margin confirms that statistical verification can provide the highly accurate, quantitative safety assurances demanded by mission-critical domains, effectively bridging the gap between exact formal proofs and the practical scalability required for real-world deployment.

\subsection{Methodological Limitations and Distributional Assumptions}
\label{secIID:Limitation}

The primary limitation of the RoMA methodology stems from its foundational reliance on the assumption that a neural network's runner-up confidence scores distribute normally under perturbation. To enforce and verify this critical statistical property, the framework utilizes the Anderson-Darling goodness-of-fit test, occasionally coupled with a Box-Cox power transformation to normalize skewed data. However, this distributional assumption is not universally satisfied across all operational domains or input modalities. 

For example, when evaluating large language models under orthographic perturbations-such as substituting single characters randomly within a sentence to simulate human typographical errors, the resulting confidence distributions frequently fail the Anderson-Darling test, even post-transformation. In instances where the data does not distribute normally, the formal statistical guarantees of the method are compromised~\cite{LeAsKa25}. Interestingly, empirical evaluations demonstrate that the framework can still yield highly accurate estimates even when normality assumptions are violated. In an exhaustive brute-force evaluation comparing RoMA's estimations of character-level noise against the true, exact distribution, the statistical robustness probabilities deviated from the ground-truth scores by merely 0.17\% (yielding estimates of 94.44\% and 93.94\% against ground truths of 94.61\% and 93.84\%, respectively)~\cite{LeAsKa25}. 

Despite this demonstrated empirical accuracy, the absence of a verified normal distribution means that formal safety assurances cannot be definitively certified for regulatory compliance. To mitigate this limitation in safety-critical contexts, an alternative strategy is to redefine the perturbation boundaries to find a closely adjacent, or ``closed'' use case where the outputs do successfully normalize. By shifting the definition of the adversarial input space-for example, transitioning from character-level noise to semantic embedding substitutions where normality is maintained in the vast majority of cases-practitioners can recover the required distributional properties and re-establish the formal statistical bounds necessary for system certification.

\section{Proposed Framework}
\label{sec4:Framework}

\subsection{From Technical Tool to Regulatory Mechanism}

The preceding sections establish two independent findings: 
\begin{inparaenum}[(i)]
\item Shows that major regulatory frameworks adopt risk-based classification without specifying what ``acceptable risk'' means in measurable terms or how compliance with such a threshold can be technically verified~\cite{euaiact2024,ebers2024}. 
\item Shows that RoMA and gRoMA can measure, with quantifiable statistical confidence, the probability that a deployed neural network produces a failure-class output within a defined perturbation domain. 
\end{inparaenum}

This section proposes how the two halves connect: a two-stage pre-deployment certification procedure in which the normative and technical tasks are formally separated and sequentially ordered.

The design responds to a structural gap in current AI regulation: existing frameworks require pre-deployment conformity assessment~\cite{euaiact2024,nist2023rmf} but provide no technical method for conducting one on statistical learning systems. Traditional software certification --- as practised under DO-178C~\cite{LaNi11} for avionics and IEC~62304 for medical device software~\cite{Jo06} --- relies on structural inspection: source code review, control-flow analysis, and formal proofs of functional correctness. These methods presuppose access to, and interpretability of, the artefact under test; neither presupposition holds for a neural network whose safety-relevant behaviour is distributed across learned weight matrices that resist structural analysis. The procedure proposed here replaces structural inspection with a sampling-based acceptance test: given a formally specified failure rate~$\delta$ and perturbation domain~$\varepsilon$, Stage Two determines, with bounded statistical error, whether the deployed system meets its safety specification --- producing a falsifiable, reproducible certificate that occupies the same logical role as a traditional conformity assessment, without requiring access to model internals.

\subsection{Stage One: The Normative Determination}

Before any technical verification can proceed, the competent authority must fix two parameters. The first is the acceptable failure probability~$\delta$: the maximum rate at which the system may produce outputs outside its certified behavior class across the operational domain. The second is the perturbation domain~$\varepsilon$, defining the input neighborhood within which the system must maintain stable behavior: the set of variations in formatting, phrasing, or sensory noise that real-world deployment will generate and within which the system's response must remain consistent.

Neither parameter can be derived from engineering considerations alone. $\delta$ is a normative decision analogous to the aviation industry's $10^{-9}$ catastrophic failure rate per flight hour - a threshold arrived at through decades of expert deliberation and public commitment, not through technical optimization~\cite{LaNi11}.

The road traffic speed limit offers a parallel from a different domain: speed limits are set not because every driver within them is guaranteed safe, but because society has implicitly accepted the residual accident rate produced by that threshold as tolerable given competing interests in mobility~\cite{kaminski2023}. Setting $\delta$ for an AI system is the same kind of determination - explicit rather than implicit, and therefore more honest.
$\varepsilon$ is a deployment-context determination requiring joint input from domain regulators and system developers, reflecting the foreseeable range of inputs the system will encounter in practice. Fixing both parameters publicly and in advance imposes a risk-communication discipline currently absent from AI governance~\cite{kaminski2023}: regulators, developers, and end users know exactly what level of residual risk is being certified, rather than relying on the process proxies - oversight logs, appointed officers, procedural safeguards - that currently substitute for outcome accountability~\cite{nist2023rmf}.
The system is certified if and only if this upper bound does not exceed $\delta$. The sample size $n$ required to achieve a given confidence level is computable in advance, making the evidentiary burden transparent and auditable. For systems whose operational domain spans multiple input categories, gRoMA (Section~\ref{seIIB:Scaling}) extends this procedure to a global categorical robustness score, aggregating per-category estimates weighted by the operational profile.

When the Anderson-Darling test rejects normality, two fallback pathways are available, as detailed in Section~\ref{secIID:Limitation}:
\begin{inparaenum}[(i)]
\item Domain narrowing redefines $\varepsilon$ to identify sub-domains where the distribution recovers approximate Gaussianity, enabling partial certification within those boundaries while explicitly flagging uncertified regions. 
\item Brute-force evaluation applies exhaustive exact counting within a bounded input space, avoiding distributional assumptions entirely but at exponential computational cost. 
\end{inparaenum}

The choice between pathways is governed by the risk profile of the deployment context: high-risk applications may require full coverage, or an explicit and documented acknowledgment that certain input sub-domains fall outside the certified scope.

\subsection{Integration with Existing Regulatory Obligations}

The proposed procedure is designed to satisfy, not displace, existing certification pipeline requirements. Under the EU AI Act, high-risk systems must demonstrate technical robustness and accuracy before market entry~\cite{euaiact2024}. RoMA and gRoMA frameworks provide the quantitative technical backbone for that demonstration, supplying auditable numerical outputs where the Act currently requires process documentation without specifying what that documentation must prove~\cite{ebers2024}. The procedure also addresses the self-certification problem that legal scholars have identified as the Act's central enforcement weakness~\cite{wachter2024}: a RoMA certificate, produced against publicly stated parameters and independently verifiable from the sampling protocol, constitutes third-party-auditable evidence of compliance in a form that lobbying-induced carve-outs cannot easily dilute.

The frameworks also address the practical inadequacy of the risk management strategies that currently substitute for quantified certification. Contemporary deployment practice places the burden of safety either on end users, through contractual disclaimers and usage guidelines, or on human-in-the-loop oversight mechanisms. Empirical research consistently shows that human oversight of high-throughput automated decision systems is limited in scope and prone to automation bias~\cite{wachter2024}. A pre-deployment statistical certificate shifts the locus of accountability upstream to the developer, where it can be independently verified, rather than downstream to operators and users who lack the technical means to assess what they are overseeing.

Under the NIST AI Risk Management Framework~\cite{nist2023rmf}, the procedure maps directly onto the Measure function, operationalizing the ``trustworthiness characteristics'' that the RMF identifies but does not quantify. The Council of Europe Framework Convention~\cite{coe2024}, which requires measures ``graduated and differentiated'' by severity and probability of harm, is similarly served: the two-stage structure separates the normative graduation ($\delta$-setting) from the technical differentiation (RoMA verification), giving each element of the treaty obligation a distinct and auditable implementation step.

Re-certification is required whenever a system is materially updated or its operational domain changes - consistent with the post-market monitoring provisions of both the AI Act and the NIST RMF. Because the RoMA procedure is black-box and automated, re-certification does not require access to model weights or architectural documentation; it requires only that the updated system be re-sampled against the same $\varepsilon$ and evaluated against the same $\delta$. This property makes the framework practically deployable by regulatory bodies that lack the internal technical capacity to audit model internals, and by third-party conformity assessment bodies operating under Article~$33$ of the AI Act~\cite{euaiact2024}.

\subsection{Scope: Safety, Not Transparency}

Unlike risk management processes, which are process-oriented and reactively applied, or post-market monitoring obligations, which are corrective, the certificate produced by Stage Two is an ex-ante safety guarantee: a mathematically bounded claim, established before deployment, that failure probability over the certified operational domain does not exceed~$\delta$. This positions the framework within the acceptance testing tradition of safety-critical systems engineering --- analogous in logical structure to qualification evidence required under DO-178C~\cite{LaNi11} and IEC~62304~\cite{Jo06}, but generalised to the statistical behaviour of learned models. The guarantee is explicitly conditional on the operational envelope specified by~$\varepsilon$: inputs outside the certified domain fall outside the certificate's scope and require runtime boundary detection to flag, a function addressed in the Operational Assurance subsection below.

The framework certifies distributional robustness within a defined input domain, bounding failure probability against a pre-specified $\delta$ without requiring access to model internals. It does not address explainability, fairness, or out-of-distribution behaviour --- these remain open problems requiring separate regulatory instruments. Transparency and explainability obligations under applicable law are addressed by complementary mechanisms outside the scope of this work.

\subsection{Legal Implications: Certified Risk as a Defense Standard}
\label{subsec:LegalImplications}

Certifying a system against a formally specified $\delta$ has direct implications for how regulatory compliance evidence is structured. The EU AI Act's Article~9 (risk management systems) and Article~15 (accuracy, robustness, and cybersecurity) both require technical documentation establishing that high-risk AI systems satisfy their requirements, but neither article specifies what form that evidence must take~\cite{euaiact2024,ebers2024}. A RoMA certificate satisfies this evidentiary role in a form particularly suited to independent oversight: it is reproducible from its publicly stated sampling protocol, auditable by third-party conformity assessment bodies operating under Article~33, and statistically bounded in a way that makes it falsifiable --- a re-test of the deployed system against the same $\delta$ and $\varepsilon$ either confirms or defeats the original certificate. This distinguishes it from the process documentation that currently substitutes for quantitative compliance evidence, which makes no falsifiable claim about system behaviour. Once such thresholds are in place, the certified failure rate may also constitute a legally cognizable benchmark analogous to risk acceptance standards in aviation and pharmaceuticals~\cite{LaNi11}; a detailed treatment of the resulting liability implications is beyond the scope of this paper.

The evidentiary value of the certificate presupposes that $\delta$ was itself determined through a legitimate normative process. The framework does not specify how that process must be conducted --- this is a governance question, not an engineering one --- but it imposes a structural precondition: a certificate is only as meaningful as the threshold it certifies against, which must have been publicly stated and arrived at through a process that is defensible to those affected by the system's deployment. This precondition operationalizes obligations that existing regulatory instruments already impose but leave technically unspecified: Article~9 of the EU AI Act requires deployers of high-risk systems to establish risk management processes that identify acceptable levels of residual risk~\cite{euaiact2024,ebers2024}, while the NIST AI RMF requires that risk tolerances be defined within the Govern function before technical measurement results can be meaningfully interpreted~\cite{nist2023rmf}. The $\delta$-setting stage thus transforms these existing but unquantified obligations into a formally specified, technically precise input that the certification procedure can act on. The engineering contribution is not to resolve the normative question but to make its resolution a hard precondition of deployment --- ensuring that the question of acceptable failure rate is asked, answered, and recorded before the system goes live.

\subsection{The Ethical Imperative: Pre-Commitment and Public Deliberation}

The normative determination of $\delta$ is a deliberate design constraint, not a limitation: the framework cannot be applied without first making explicit, through legitimate deliberative processes, what failure rate society accepts for a given deployment context. This forces value judgments that are typically absorbed silently into system design into a separate, auditable, and contestable pre-deployment stage. The framework thus operationalizes a form of accountability-by-architecture --- robustness certification is only meaningful once the moral debate about what counts as acceptable risk has been resolved publicly. In requirements engineering terms, $\delta$ functions as an acceptance criterion: it cannot be inferred from system behaviour, derived from training data, or delegated to the developer --- it must be stipulated externally before Stage Two can have a determinate pass/fail interpretation. This discipline is structurally analogous to safety integrity level assignment under IEC~61508, where the level must be fixed prior to system design because it determines what evidence of compliance is required. In practice, the $\delta$ value and its justification become part of the formal certification record --- auditable by regulators, contestable by affected parties, and revisable as deployment contexts change --- externalizing a value judgment that would otherwise be embedded invisibly in the developer's choice of training objective or deployment threshold.

\subsection{Operational Assurance: From Static Certificate to Runtime Monitoring}

The statistical certificate produced in Stage Two is not merely a static regulatory artifact; it serves as a critical parameter for \emph{Runtime Assurance (RA)} architectures. In a safety-critical system, the certified failure bound $\delta$ provides the baseline for an onboard \emph{safety monitor}. If the environmental conditions shift beyond the defined perturbation domain $\varepsilon$---for example, due to extreme weather conditions not covered in the original certificate---the runtime monitor can trigger a fallback to a \emph{Minimum Risk Maneuver (MRM)}~\cite{BrPa23,KaLeRaYe24}.

Furthermore, the framework addresses the \emph{Confidence-Sample Trade-off}. To certify a safety-critical threshold of $\delta = 10^{-9}$ with a confidence level of $1-\alpha = 0.99$, Hoeffding’s inequality allows the developer to pre-calculate the required sample size $n$. This transparency in the evidentiary burden is vital for resource planning in high-resolution vision systems, ensuring that the certification process itself is as thorough and predictable as the airborne software standards from which it draws inspiration.

\section{Case Study: Certifying an Autonomous Emergency Braking Vision System}
\label{sec5:caseStudy}

To illustrate the practical deployment of the proposed framework, we examine a safety-critical Autonomous Emergency Braking (AEB) system~\cite{YaYaWu2022}. Modern AEB architectures rely on Deep Neural Networks for sensor fusion to initiate stopping protocols~\cite{FuLiYu20}. As a high-risk system, it demands quantifiable safety assurances before public deployment.

\subsection{Stage One: Normative Parameterization and Risk Budgeting}
\label{subsec:StageOneCaseStudy}

The process begins with the regulator and developer defining operational boundaries. We adopt the aerospace baseline failure probability of $\delta = 10^{-9}$ per operational hour~\cite{LaNi11}. To operationalize this, we propose a \emph{Risk Budgeting} approach, where $\delta$ is treated as a systemic budget decomposed across $k$ distinct failure modes within the Operational Design Domain (ODD). 

The aggregated risk $P_{total}$ is defined by the following exposure-weighted summation:
\begin{equation}
P_{total} \approx \sum_{i=1}^{k} \omega_i \cdot P(\text{failure} \mid \varepsilon_i) \leq \delta
\end{equation}
where $\varepsilon_i$ represents a specific perturbation sub-domain and $\omega_i$ denotes the \emph{exposure weight}---the estimated frequency of encountering that condition during deployment.

\subsubsection{Failure Mode Decomposition}
Following the principles of Failure Mode and Effects Analysis (FMEA), we decompose the AEB sensor's attack surface into three measurable sub-domains:
\begin{itemize}
    \item \emph{Transient Interference ($\varepsilon_{glare}$):} Simulating sensor saturation from sunlight or high-beams ($\omega_{high}$ but transient).
    \item \emph{Optical Degradation ($\varepsilon_{scratches}$):} Simulating physical lens damage or occlusion. While the failure probability per encounter is high, the exposure rate is lower ($\omega_{low}$).
    \item \emph{Stochastic Noise ($\varepsilon_{thermal}$):} Simulating baseline Gaussian noise from sensor aging, constant across all operational hours ($\omega = 1$).
\end{itemize}
By defining specific risk budgets $\delta_i$ for each mode such that $\sum \omega_i \delta_i \leq \delta$, the regulator transforms a vague safety goal into a definitive engineering checklist.

\subsection{Stage Two: Executing Statistical Verification} 

With thresholds established, the developer initiates the gRoMA protocol to evaluate the target category $c$ (``pedestrian detected''). The procedure samples $n$ correctly classified images from the operational domain. For each image $x_i$, RoMA generates perturbed inputs within the $\varepsilon$-neighborhood, simulates environmental variance, and extracts runner-up confidence scores.

We utilize the Anderson-Darling test and Box-Cox transformations to ensure the distributions satisfy normality. Once validated, the algorithm calculates $p_i$, the empirical probability that a perturbation within $\varepsilon$ causes a misclassification. Finally, gRoMA aggregates these local metrics and applies Hoeffding's inequality to compute a definitive upper bound on the global failure rate $P$. The model achieves certification if and only if $P \leq \delta$.

\subsection{Certification Outcome and Lifecycle Agility}

If $P \leq 10^{-9}$, the model is granted a formal statistical certificate. Conversely, failure to meet this threshold allows developers to isolate specific perturbation manifolds (e.g., specific angles of glare) for targeted retraining. This creates an iterative recertification loop that aligns with industrial CI/CD (Continuous Integration/Continuous Deployment) pipelines, ensuring quantifiable safety without the administrative delays of traditional software recertification.

\subsection{From Academic Benchmarks to Industrial Certification}

This case study is presented as a structured proof-of-concept: it demonstrates that the two-stage framework produces a well-defined, auditable certification output for an architecture and failure class of practical relevance, and that the procedure terminates with a falsifiable pass/fail result on a high-resolution safety-critical system. Unlike foundational RoMA evaluations on low-resolution benchmarks such as CIFAR-10, the AEB scenario addresses a domain where failure carries direct physical and legal consequences, validating that the statistical properties of the framework hold under industrially realistic conditions. A full empirical validation would require certification against a real deployed system, comparison of the certified~$\delta$ against an independently established regulatory threshold, and longitudinal re-certification across successive software updates --- directions we identify as immediate priorities for future work.

\section{Threats to Validity}
\label{sec6:ThreatsToValidity}

While the proposed two-stage certification framework provides a measurable pathway for regulating high-risk artificial intelligence, its theoretical foundations and practical implementations are subject to several validity threats. Following standard software engineering and reliability research practices, we categorize these limitations into construct, internal, external, and statistical validity. Acknowledging these constraints is essential for guiding future research and preventing the over-reliance on statistical certificates in production environments.

\subsection{Construct Validity}

Construct validity concerns whether the proposed methodology accurately measures the theoretical concept of ``safety'' it intends to evaluate. The primary threat in this domain stems from the maturity of the underlying verification tools. The framework relies heavily on the adaptation of RoMA and gRoMA. Currently, these are state-of-the-art academic research methodologies; they have not yet been adopted by the broader industry nor subjected to the extensive, decades-long stress testing that traditional software standards like DO-178 have undergone. Consequently, while the mathematical translation of aviation fault-tolerance paradigms to deep neural networks is theoretically sound, the practical friction of integrating these tools into existing industrial CI/CD pipelines remains largely unmapped. 

Furthermore, defining the acceptable failure probability ($\delta$) and the perturbation domain ($\varepsilon$) involves subjective regulatory judgment. There is a risk that an improperly defined $\varepsilon$ might not accurately reflect the true latent hazards of the deployment environment, leading to a certificate that ensures compliance against a flawed metric rather than ensuring actual operational safety.

\subsection{Internal Validity}

Internal validity examines the foundational assumptions and internal mechanics of the experimental or methodological design. The most critical threat to this framework is its strict dependence on the normal distribution of the neural network's runner-up confidence scores. The RoMA methodology assumes that local perturbations will yield normally distributed outputs, verified via the Anderson-Darling test. 

While the algorithm incorporates a Box-Cox power transformation to correct skewed data, empirical evidence indicates that in certain complex topologies or specific data modalities, the data remains resolutely non-normal even post-transformation~\cite{LeAsKa25}. If the normality assumption is fundamentally violated and no adjacent operational domain can be found to recover it, the mathematical proofs underpinning the framework break down. In such scenarios, the tool cannot provide a valid safety certificate, leaving exhaustive, computationally prohibitive formal verification as the only fallback.

Additionally, as a statistical methodology, the framework is intrinsically vulnerable to sampling constraints. The integrity of the local robustness score dictates that the algorithm must explore a highly representative subset of the $\varepsilon$-neighborhood. If the random sampling process fails to uncover a narrow but catastrophic vulnerability manifold---a common anomaly in highly non-linear, over-parameterized models---the resulting robustness probability will be artificially inflated, falsely certifying an unsafe system.

\subsection{External Validity}

External validity addresses the generalizability of the certification framework across different operational conditions and data modalities. The statistical bounds calculated during Stage Two are strictly confined to the agreed-upon operational domain and the predefined perturbation space ($\varepsilon$). The framework does not guarantee safety against Out-of-Distribution (OOD) inputs, structural data drift, or novel ``black swan'' events that the model has never encountered. If the real-world deployment environment shifts materially from the parameters tested during certification, the certified failure rate ($\delta$) is immediately invalidated.

Moreover, while the framework translates exceptionally well to continuous input spaces such as optical sensors (e.g., the autonomous braking case study), its generalizability to discrete input spaces, such as Natural Language Processing (NLP) and Large Language Models (LLMs), presents distinct challenges. Small syntactic perturbations in text do not behave identically to Gaussian noise in vision models~\cite{LeAsKa25}, often requiring complex semantic embedding adaptations to fulfill the algorithm's distributional prerequisites.

\subsection{Statistical and Conclusion Validity}

Statistical validity pertains to the mathematical power and limitations of the quantitative claims. Unlike formal verification methods using Satisfiability Modulo Theories (SMT) solvers, which provide deterministic, absolute proofs of network properties, this framework provides a probabilistic approximation. It inherits all the fundamental challenges associated with statistical estimation.

The framework utilizes Hoeffding's inequality to establish a definitive upper bound on the failure probability. While this guarantees that the error margin is strictly controlled by the selected confidence level ($1-\alpha$), Hoeffding's bounds are notoriously conservative. In practice, this means the framework might drastically overestimate the true failure probability to maintain mathematical safety guarantees. Consequently, developers might be forced to dedicate substantial resources to retrain or abandon models that are, in reality, sufficiently safe but cannot be certified due to the conservative nature of the statistical bound. 

Finally, there always exists a non-zero probability ($\alpha$) that the true failure rate exceeds the certified $\delta$ threshold, a foundational reality of probabilistic safety that regulators and legal frameworks must explicitly accommodate.

\section{Conclusion and Future Research Directions}
\label{sec7:conclusion}

\subsection{Closing the Loop}

This paper began from a structural observation: risk-based AI regulation has become the dominant global governance model, yet no major framework specifies what ``acceptable risk'' means in quantitative terms or provides a technical method for verifying that a deployed system actually meets such a threshold. The regulatory architecture exists; the measuring instrument has been missing.

We have proposed that instrument. The two-stage certification framework presented in Section~\ref{sec4:Framework} - in which a competent authority first fixes an acceptable failure probability~$\delta$ and a perturbation domain~$\varepsilon$, and a developer then verifies compliance using the RoMA and gRoMA statistical tools, transforms an open-ended process obligation into a falsifiable outcome test. The case study in Section~\ref{sec5:caseStudy} demonstrates that the procedure is not merely theoretical: applied to an Autonomous Emergency Braking system, it produces a concrete, auditable certificate with a defined scope of validity and a clear re-certification trigger. The threats identified in Section~\ref{sec6:ThreatsToValidity} are real, but they are bounded: they define the conditions under which the framework applies, rather than undermining it wholesale.

The contribution is therefore a practical, if partial, solution. It does not resolve every challenge AI governance faces - explainability, bias, systemic concentration risk, and the governance of general-purpose models all remain open problems. What it does is eliminate the single most consequential gap: the absence of a pre-deployment, outcome-based safety certificate for neural networks operating in high-risk domains.

\subsection{Future Technical Work}

The most immediate technical priority is a systematic mapping of the framework's applicability across input modalities and model architectures. RoMA's normality assumption holds reliably for image classifiers and structured-input networks, but fails for large language models under orthographic perturbation, as Section~\ref{secIID:CommonAssumption} documents. Future work should characterize precisely which perturbation types and model families satisfy the assumption, producing a publicly accessible applicability taxonomy that practitioners and regulators can consult before initiating certification. Equally important is the complementary mapping of \emph{inapplicable} cases, not to foreclose regulation, but to define the frontier where enhanced human-in-the-loop mechanisms and provisional safeguards remain the appropriate interim response, and to direct research effort toward closing those gaps.

A second technical priority is computational scaling for large language models. The brute-force fallback described in Section~\ref{secIID:CommonAssumption} is currently intractable for networks exceeding a few thousand neurons. Research into more efficient distributional testing methods - including conformal prediction approaches and distribution-free concentration inequalities - could extend certification to architectures that currently lie beyond the method's reach.

\subsection{Future Regulatory Work}

The normative stage of the proposed framework --- Stage~One, the public determination of~$\delta$ --- cannot be accomplished by engineers or left to self-certification. It requires deliberate regulatory action, and that action has not yet been taken in any major jurisdiction. We call for the initiation of structured public deliberation on acceptable failure rates for high-risk AI categories, analogous to the expert consensus processes that produced the $10^{-9}$ standard in aviation and the statistical tolerances embedded in pharmaceutical approval regimes.

This deliberation should take account of a specific challenge that AI introduces but prior industries did not face at scale: AI systems are increasingly deployed in situations that previously demanded human instinct rather than calculated risk - emergency response, real-time collision avoidance, acute clinical triage. In these contexts, there is no pre-existing societal consensus on acceptable failure rates, and no historical accident record from which to derive one. The~$\delta$-setting process must therefore generate that consensus, not merely encode existing norms.

International standardization bodies --- IEEE, ISO, and EUROCAE among them --- are the appropriate institutional venues for translating agreed~$\delta$ values into enforceable technical standards. The current generation of AI standards is almost exclusively process-based. The framework proposed here provides a template for standards that are outcome-based: specifying not how a system must be built or governed, but what it must demonstrably be able to do before it is deployed.

\subsection{A Call for Interdisciplinary Research}

The framework presented in this paper sits at the boundary between computer science and law. Its technical components --- the RoMA algorithm, the Hoeffding bound, the Anderson-Darling test --- are products of the formal methods and machine learning literatures. Its normative components --- the choice of~$\delta$, the definition of high-risk categories, the legal effect of certification --- are products of regulatory theory and legislative design. Neither discipline can complete the project alone.

We call for sustained, institutionalized collaboration between these communities: joint working groups, shared publication venues, and regulatory sandboxes in which certification procedures can be piloted on real systems before being enshrined in binding law. The gap between what AI can do and what governance frameworks can verify has widened for a decade. The tools to begin closing it now exist. Whether they are deployed depends not on further technical invention, but on the political will to demand measurable outcomes from AI regulation, and the scientific community's willingness to supply the methods that make such demands enforceable.

\addcontentsline{toc}{section}{References}

\section*{References}
\printbibliography[heading=none]

@article{An11,
	author = {Anderson, Theodore W.},
	journal = {Int. Encyclopedia of Statistical Science},
	volume = {1},
	pages = {52--54},
	title = {{Anderson-Darling Tests of Goodness-of-Fit}},
	year = {2011}
}

@article{BoCo82,
	author = {Box, George Edward and Cox, David },
	journal = {Journal of the American Statistical Association},
	number = {377},
	pages = {209-210},
	title = {{An Analysis of Transformations Revisited, Rebutted}},
	volume = {77},
	year = {1982}
}

@article{ho63,
  title={{Probability Inequalities for Sums of Bounded Random Variables}},
  author={Hoeffding, Wassily},
  journal={Journal of the American statistical association},
  volume={58},
  number={301},
  pages={13--30},
  year={1963},
  publisher={Taylor \& Francis}
}

@article{KaBaDiJuKo21,
 author={Katz, Guy and Barrett, Clark and Dill, David L and Julian, Kyle and Kochenderfer, Mykel J},
 title = {{Reluplex: a Calculus for Reasoning about Deep Neural Networks}},
 journal = {Formal Methods in System Design (FMSD)},
 year = {2021}
}

@inproceedings{CoRoZi19,
  title={{Certified Adversarial Robustness via Randomized Smoothing}},
    author={Cohen, Jeremy and Rosenfeld, Elan and Kolter, Zico},
  booktitle={Proc. 36th Int. Conf. on Machine Learning (ICML)},
  year={2019}
}

@article{LaNi11,
  title={{ARP4754A/ED-79A-Guidelines for Development of Civil Aircraft and Systems-Enhancements, Novelties and Key Topics}},
  author={Landi, A. and Nicholson, M.},
  journal={SAE Int. Journal of Aerospace},
  volume={4},
  pages={871--879},
  year={2011}
}

@misc{FAA93,
author = {{Federal Aviation Administration}},
title = {{ RTCA, Inc., Document RTCA/DO-178B }},
year = { 1993 },
note = { \url{https://nla.gov.au/nla.cat-vn4510326} }
}

@inproceedings{LeKa21,
  title = {{RoMA: a Method for Neural Network Robustness Measurement and Assessment}},
  author = {Levy, Natan and Katz, Gutz},
  booktitle = {Proc. 29th Int. Conf. on Neural Information Processing (ICONIP)},
  Year = {2021}
}

@misc{euaiact2024,
  author       = {{European Parliament and Council of the European Union}},
  title        = {Regulation ({EU}) 2024/1689 Laying Down Harmonised Rules on Artificial Intelligence ({Artificial Intelligence Act})},
  year         = {2024},
  howpublished = {Official Journal of the European Union},
  note         = {OJ L 2024/1689, 12 July 2024}
}

@misc{ec2025gpai,
  author       = {{European Commission}},
  title        = {Guidelines on the Application of {Article~55} of {Regulation (EU) 2024/1689} to General-Purpose {AI} Models},
  year         = {2025},
  howpublished = {European Commission Staff Document}
}

@article{ebers2024,
  author  = {Ebers, Martin},
  title   = {Truly Risk-Based Regulation of {AI}},
  journal = {European Journal of Risk Regulation},
  year    = {2024},
  note    = {Advance publication}
}

@article{wachter2024,
  author  = {Wachter, Sandra},
  title   = {The {EU} {AI} Act: A {Political} Not a Risk-Based Approach},
  journal = {Yale Journal of Law and Technology},
  volume  = {26},
  year    = {2024}
}

@article{laux2024,
  author  = {Laux, Johann and Wachter, Sandra and Mittelstadt, Brent},
  title   = {Trustworthy Artificial Intelligence and the {European Union} {AI} Act: On the Conflation of Trustworthiness and Conformity with Ethical Principles},
  journal = {Regulation \& Governance},
  volume  = {18},
  number  = {1},
  year    = {2024}
}

@misc{cac2022algo,
  author       = {{Cyberspace Administration of China}},
  title        = {{Provisions on the Management of Algorithmic Recommendations in Internet Information Services
                  }},
  year         = {2022},
  howpublished = {Order No.~9 of the Cyberspace Administration of China},
  note         = {Effective 1 March 2022}
}

@misc{cac2023synthesis,
  author       = {{Cyberspace Administration of China}},
  title        = {{Provisions on the Management of Deep Synthesis Internet Information Services
                  }},
  year         = {2023},
  howpublished = {Cyberspace Administration of China},
  note         = {Effective 10 January 2023}
}

@misc{cac2023genai,
  author       = {{Cyberspace Administration of China}},
  title        = {Interim Measures for the Management of Generative Artificial Intelligence Services
                  },
  year         = {2023},
  howpublished = {Cyberspace Administration of China},
  note         = {Effective 15 August 2023}
}

@techreport{nist2023rmf,
  author      = {{National Institute of Standards and Technology}},
  title       = {Artificial Intelligence Risk Management Framework ({AI RMF} 1.0)},
  number      = {NIST AI 100-1},
  institution = {U.S. Department of Commerce, National Institute of Standards and Technology},
  year        = {2023},
  doi         = {10.6028/NIST.AI.100-1}
}

@techreport{nist2024genai,
  author      = {{National Institute of Standards and Technology}},
  title       = {{Artificial Intelligence Risk Management Framework: Generative Artificial Intelligence Profile}},
  number      = {NIST AI 600-1},
  institution = {U.S. Department of Commerce, National Institute of Standards and Technology},
  year        = {2024},
  doi         = {10.6028/NIST.AI.600-1}
}

@misc{trump2025eo,
  author       = {{Executive Office of the President}},
  title        = {{Executive Order 14179: Removing Barriers to {American} Leadership in Artificial Intelligence}},
  year         = {2025},
  howpublished = {Federal Register},
  volume       = {90},
  number       = {14},
  note         = {Signed 23 January 2025, 90 Fed.\ Reg.\ 8741}
}

@article{stanovsky2025,
  author  = {Stanovsky, Gabriel and Keydar, Renana and Perl, Gadi and Habba, Tamar},
  title   = {{Beyond Benchmarks: On the False Promise of {AI} Regulation}},
  journal = {AI Regulation},
  volume  = {6},
  year    = {2025}
}

@misc{coe2024,
  author       = {{Council of Europe}},
  title        = {{Framework Convention on Artificial Intelligence and Human Rights, Democracy and the Rule of Law}},
  year         = {2024},
  howpublished = {Council of Europe Treaty Series No.~225 (CETS No.~225)},
  note         = {Opened for signature 5 September 2024; entered into force 1 November 2025}
}

@misc{eu2025omnibus,
  author       = {{European Commission}},
  title        = {{Proposal for a Regulation of the {European Parliament} and of the {Council} Amending Regulations on Digital Legislation ({Digital Omnibus})}},
  year         = {2025},
  howpublished = {COM(2025) 836 final}
}

@article{kaminski2023,
  author  = {Kaminski, Margot E.},
  title   = {{Binary Governance: Lessons from the {EU} {AI} Act's Risk Classification of {AI} Systems}},
  journal = {Boston University Law Review},
  volume  = {103},
  number  = {4},
  pages   = {1529--1579},
  year    = {2023}
}

@inproceedings{rushby2011new,
  title={{New Challenges in Certification for Aircraft Software}},
  author={Rushby, John},
  booktitle={Proceedings of the ninth ACM international conference on Embedded software},
  pages={211--218},
  year={2011}
}

@inproceedings{gabreau2022toward,
  title={{Toward the Certification of Safety-Related Systems Using ML Techniques: the ACAS-XU Experience}},
  author={Gabreau, Christophe and Gauffriau, Adrien and De Grancey, Florence and Ginestet, Jean-Brice and Pagetti, Claire},
  booktitle={11th European Congress on Embedded Real Time Software and Systems (ERTS 2022)},
  year={2022}
}

@article{LeYeKa23,
  title={{gRoMA: a Tool for Measuring Deep Neural Networks Global Robustness}},
  author={Levy, Natan and Yerushalmi, Raz and Katz, Guy},
  journal={arXiv preprint arXiv:2301.02288},
  year={2023}
}

@inproceedings{LeAsKa25,
  title={{Statistical Runtime Verification for LLMs via Robustness Estimation}},
  author={Levy, Natan and Ashrov, Adiel and Katz, Guy},
  booktitle={International Conference on Runtime Verification},
  pages={457--476},
  year={2025},
  organization={Springer}
}

@inproceedings{HuHuHuPe21,
	author = {Huang, C. and Hu, Z. and Huang, X. and Pei, K.},
	booktitle = {ICANN},
	pages = {79--90},
	title = {{Statistical Certification of Acceptable Robustness for Neural Networks}},
	year = {2021}
}

@misc{WeRaTeYe18,
        Note = {\url{http://arxiv.org/abs/1811.07209}},
	author = {Webb, S. and Rainforth, T. and Teh, Y. and {Pawan Kumar}, M.},
	title = {{A Statistical Approach to Assessing Neural Network Robustness}},
	year = {2018}
}

@article{YaYaWu2022,
  title={{A Systematic Review of Autonomous Emergency Braking System: Impact Factor, Technology, and Performance Evaluation}},
  author={Yang, Lan and Yang, Yipeng and Wu, Guoyuan and Zhao, Xiangmo and Fang, Shan and Liao, Xishun and Wang, Runmin and Zhang, Mengxiao},
  journal={Journal of advanced transportation},
  volume={2022},
  number={1},
  pages={1188089},
  year={2022},
  publisher={Wiley Online Library}
}

@article{FuLiYu20,
  title={{A Decision-Making Strategy for Vehicle Autonomous Braking in Emergency via Deep Reinforcement Learning}},
  author={Fu, Yuchuan and Li, Changle and Yu, Fei Richard and Luan, Tom H and Zhang, Yao},
  journal={IEEE transactions on vehicular technology},
  volume={69},
  number={6},
  pages={5876--5888},
  year={2020},
  publisher={IEEE}
}

@article{li2024china,
  author    = {Li, Barbara and Zhou, Amaya},
  title     = {{Navigating the Complexities of {AI} Regulation in China}},
  journal   = {Reed Smith Perspectives},
  year      = {2024},
  month     = {August},
  day       = {7},
  url       = {https://www.reedsmith.com/en/perspectives/2024/08/navigating-the-complexities-of-ai-regulation-in-china},
  note      = {Reed Smith In-depth, 2024-166},
  urldate   = {2024-11-20}}

@misc{BrPa23,
  title={{Runtime Assurance of Aeronautical Products: Preliminary Recommendations}},
  author={Brat, Guillaume and Pai, Ganeshmadhav},
  year={2023}
}

@inproceedings{KaLeRaYe24,
  title={{DEM: a Method for Certifying Deep Neural Network Classifier Outputs in Aerospace}},
  author={Katz, Guy and Levy, Natan and Refaeli, Idan and Yerushalmi, Raz},
  booktitle={2024 AIAA DATC/IEEE 43rd Digital Avionics Systems Conference (DASC)},
  pages={1--8},
  year={2024},
  organization={IEEE}
}

@inproceedings{Jo06,
  title={{Standard IEC 62304-Medical Device Software Lifecycle Processes}},
  author={Jordan, Peter},
  booktitle={2006 IET Seminar on Software for Medical devices},
  pages={41--47},
  year={2006},
  organization={IET}
}
\end{document}